\pdfoutput=1

\documentclass[11pt]{article}

\usepackage[final]{acl}

\usepackage{times}
\usepackage{latexsym}
\usepackage{booktabs}
\usepackage{color, soul}
\usepackage[T1]{fontenc}

\usepackage[utf8]{inputenc}
\usepackage[inkscapelatex=false]{svg}
\usepackage{microtype}

\usepackage{inconsolata}
\usepackage{multirow, makecell}

\usepackage{graphicx}
\usepackage{amsmath}

\usepackage{graphicx}
\usepackage{subcaption}

\sethlcolor{pink}
\title{GenEOL: Harnessing the Generative Power of LLMs for Training-Free Sentence Embeddings}

\author{
 \textbf{Raghuveer Thirukovalluru\textsuperscript{1}},
 \textbf{Bhuwan Dhingra\textsuperscript{1}},
\\
\\
 \textsuperscript{1}Duke University,
\\
\texttt{raghuveer.thirukovalluru@duke.edu} 
 }

\newcommand{\ours}{\texttt{GenEOL}}
\newcommand{\sumcse}{\texttt{SumCSE}}
\newcommand{\syncse}{\texttt{SynCSE}}
\newcommand{\geneol}{\texttt{GenEOL}}
\newcommand{\genseol}{\texttt{Gen*EOL}}
\newcommand{\seol}{\texttt{*EOL}}
\newcommand{\echo}{\texttt{Echo}}
\newcommand{\metaeol}{\texttt{MetaEOL}}
\newcommand{\keeol}{\texttt{KEEOL}}
\newcommand{\tseol}{\texttt{TSEOL}}
\newcommand{\gentseol}{\texttt{GenTSEOL}}
\newcommand{\prompteol}{\texttt{PromptEOL}}
\newcommand{\pcoteol}{\texttt{PCoTEOL}}

\newcommand{\lmthree}{\texttt{Llama3-8B}}
\newcommand{\lmtwo}{\texttt{Llama2-7B}}
\newcommand{\msone}{\texttt{Mistral0.1-7B}}
\newcommand{\Avg}{\texttt{Token Avg.}}
\newcommand{\misione}{\texttt{Mistral0.1-I-7B}}
\newcommand{\chatgpt}{\texttt{ChatGPT}}
\newcommand{\llmtovec}{\texttt{LLM2Vec}}
\newcommand{\simcse}{\texttt{SimCSE}}

\begin{document}
\maketitle
\begin{abstract}
Training-free embedding methods directly leverage pretrained large language models (LLMs) to embed text, bypassing the costly and complex procedure of contrastive learning. Previous training-free embedding methods have mainly focused on optimizing embedding prompts and have overlooked the benefits of utilizing the generative abilities of LLMs. We propose a novel method, \geneol, which uses LLMs to generate diverse transformations of a sentence that preserve its meaning, and aggregates the resulting embeddings of these transformations to enhance the overall sentence embedding. \geneol\ significantly outperforms the existing training-free embedding methods by an average of 2.85 points across several LLMs on the sentence semantic text similarity (STS) benchmark. \geneol\ also achieves notable gains in clustering, reranking, and pair-classification tasks from the MTEB benchmark. Additionally, \geneol\ stabilizes representation quality across LLM layers and remains robust to perturbations of embedding prompts.

\end{abstract}

\section{Introduction}

While LLMs are very good at generating text, the embeddings obtained from their activations are often not suitable for downstream tasks such as computing sentence similarity or clustering. 
Contrastive learning (CL) is typically used to further finetune LLMs to produce better embeddings.
Traditional contrastive learning approaches used human-annotated data for training embeddings \cite{gao2021simcse}, while recent methods leverage LLMs to generate contrastive data \cite{wang2023improving, zhang2023contrastive,thirukovalluru2024sumcse}. However, curating high-quality CL training data is both costly and time-consuming, and CL further requires large batch sizes and extensive computational resources for each round of training \cite{wang2023improving, muennighoff2024generative}. As newer language models are continuously developed and released, it is important to consider training-free, inference-time text embedding methods, which may provide a more efficient and adaptable alternative to training-intensive techniques.

\citet{jiang2023scaling} first explored this task for sentences by proposing prompts like \textit{This sentence: "[TEXT]" means in one word:"} and used the hidden layer representation of the last token from an LLM to represent \textit{[TEXT]}. \citet{zhang2024simple} enhanced the approach by incorporating chain-of-thought and knowledge reasoning. Similarly, \citet{lei2024meta} utilized multiple diverse prompts to capture different sentence aspects, averaging their embeddings for improved results. Concurrently, \citet{springer2024repetition} repeated the original sentence to facilitate bidirectional attention in embedding sentences. 

\begin{figure*}[ht]
    \centering
    \includegraphics[width=0.95\textwidth]{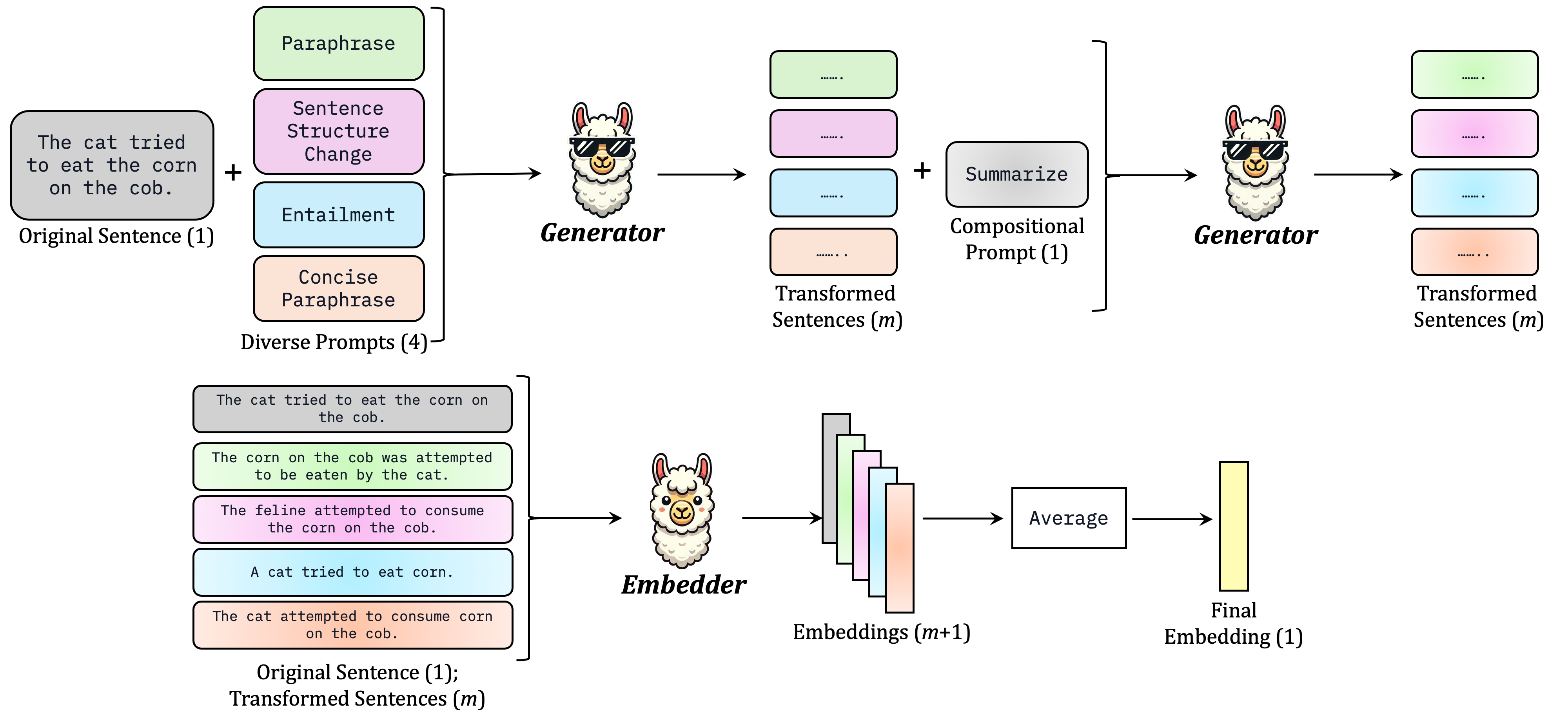}
    \caption{\geneol\ methodology outlined. Step 1: \textit{Generator} ($\mathbf{\mathcal{L}_{IT}}$) creates a set of transformed sentences, each conveying the same core meaning as the original sentence. Step 2: Original sentence, transformed sentences are embedded using \textit{Embedder} ($\mathbf{\mathcal{L}_{PT}}$) and averaged to produce the final embedding. (.) is the count of each element.
    }
    \label{fig:main}
    \vspace{-1.3em}
\end{figure*}

While these methods have significantly advanced training-free embeddings, they have not fully exploited the generative capabilities of LLMs to improve embedding quality. Recently, multiple techniques have begun scaling inference-time compute to enhance the reasoning and generative abilities of LLMs \cite{brown2024large,liang2024improving}; we adopt a similar approach for embeddings in this work. Our approach offers a method to harness the generative capabilities of large language models (LLMs), to enhance sentence embeddings. It can also work with black-box language models. Specifically, we prompt an LLM to generate $m$ diverse sentence variations that maintain the original meaning, which are then aggregated to produce more refined embeddings.


Our key contributions are as follows:
1. We show that diverse meaning retaining transformations are helpful in improving training free embeddings.
2. The proposed method, \geneol, at a higher capacity ($m$=32), significantly outperforms the previous best training-free method by (3.88, 1.83, 2.83) points on average with (\texttt{Mistral0.1-7B}, \texttt{Llama2-7B}, \texttt{Llama3-8B}) resp. on the STS benchmark \cite{conneau2018senteval}. \geneol\ with as few as two transformed sentences ($m$=2) surpasses all baselines.
3. \geneol\ outperforms other training-free methods on diverse MTEB tasks, even surpassing strong unsupervised methods (that include contrastive training) like \llmtovec\ \cite{behnamghader2024llm2vec}.
4. \geneol\ stabilizes representational quality across different LLM layers. \geneol\ is also robust to perturbations in embedding prompts.

\section{Background and Related Work}
This section covers contrastive learning (CL) and generation-based methods utilizing CL, followed by an overview of training-free approaches.\vspace{0.3em}

\noindent\textbf{CL Training}: Contrastive training employs \textit{(anchor, positive, negative)} data, where the positive is semantically similar to the anchor and the negative is dissimilar. InfoNCE loss \cite{gao2021simcse} draws the positives' embeddings closer to that of the anchor while distancing the negatives. SimCSE \cite{gao2021simcse} applies this loss using human-annotated data to train sentence embedding models.\vspace{0.3em}

\noindent\textbf{Generating data for CL Training}: E5 \cite{wang2023improving} used ChatGPT to generate a huge CL corpus of related positives and unrelated negatives across multiple tasks. Gecko \cite{lee2024gecko} used an LLM to generate queries and relabel positives, negatives from existing data corpus. \llmtovec\ \cite{behnamghader2024llm2vec} used representations generated from different dropout masks as positives with a positive only CL loss.
For sentences specifically, \syncse\ \cite{zhang2023contrastive} defined specific transformations with ChatGPT to develop contrastive positives, negatives for sentence embedding training. Inspired by compositional transformations in CL data for computer vision (CV), \sumcse\ \cite{thirukovalluru2024sumcse} further improved these transformations by using \textit{`summary'} operations (akin to cropping in CV). Our method, \ours, takes inspiration from \syncse, \sumcse\ in generating meaning retaining transformations.\vspace{0.2em}
\begin{figure*}[t]
    \centering
    \begin{subfigure}[b]{0.37\textwidth}
        \centering
         \includegraphics[width=\textwidth]{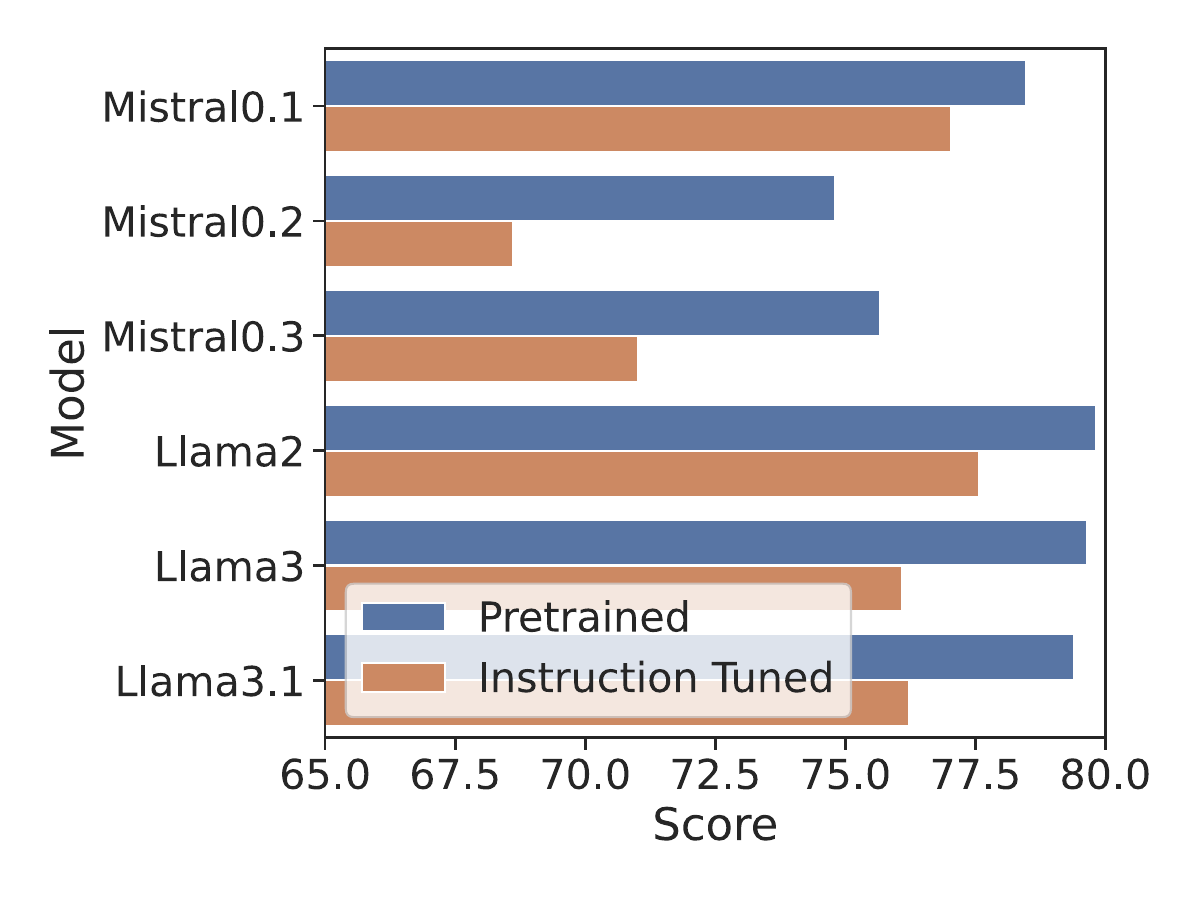}
         \caption{}
        \label{fig:obs1}
        \vspace{-0.1em}
    \end{subfigure}
    \hfill
    \begin{subfigure}[b]{0.3\textwidth}
        \centering
        \includegraphics[width=\textwidth]{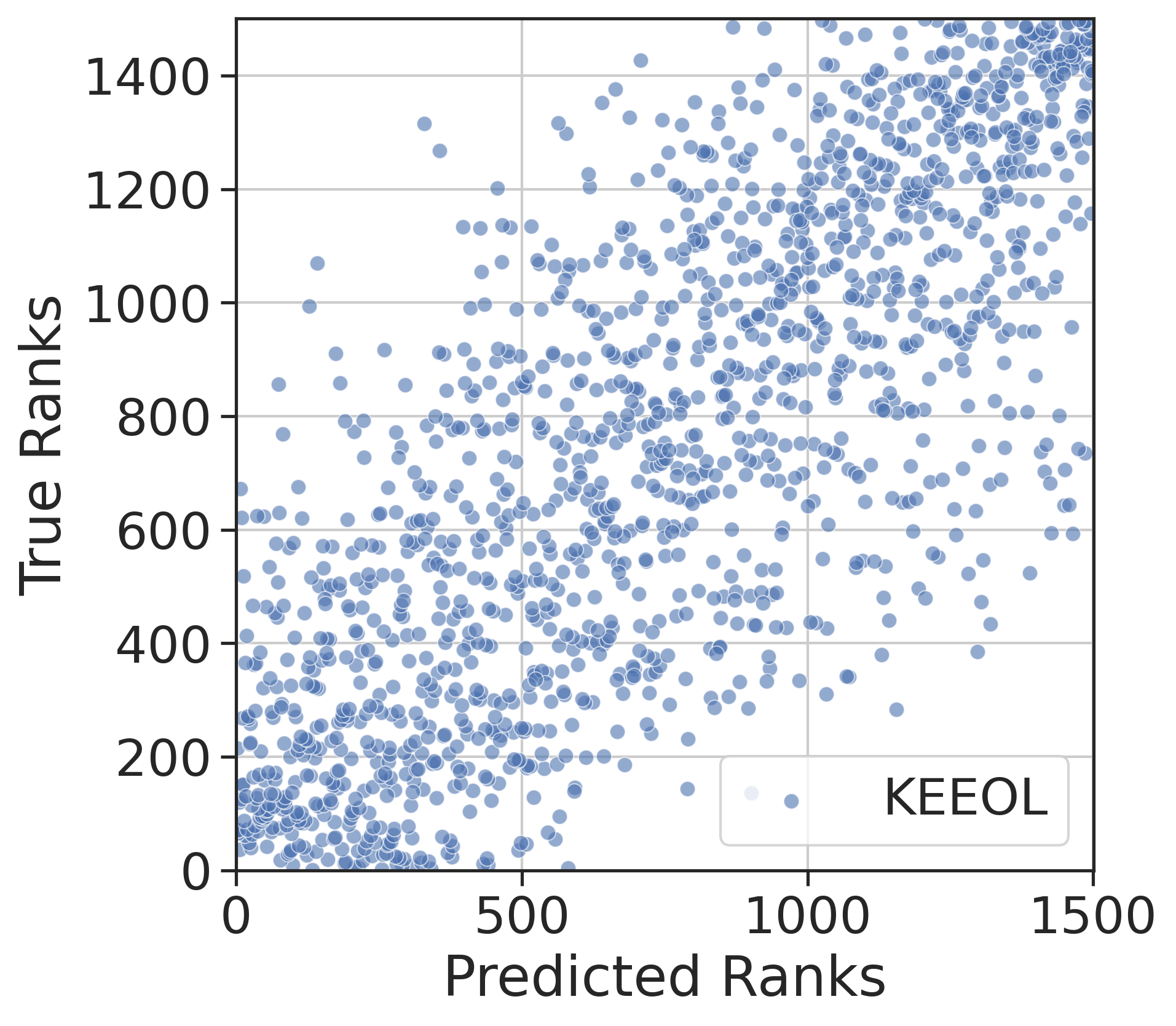}
        \caption{}
        \label{fig:obs2}
    \end{subfigure}
    \hfill
    \begin{subfigure}[b]{0.3\textwidth}
        \centering
        \includegraphics[width=\textwidth]{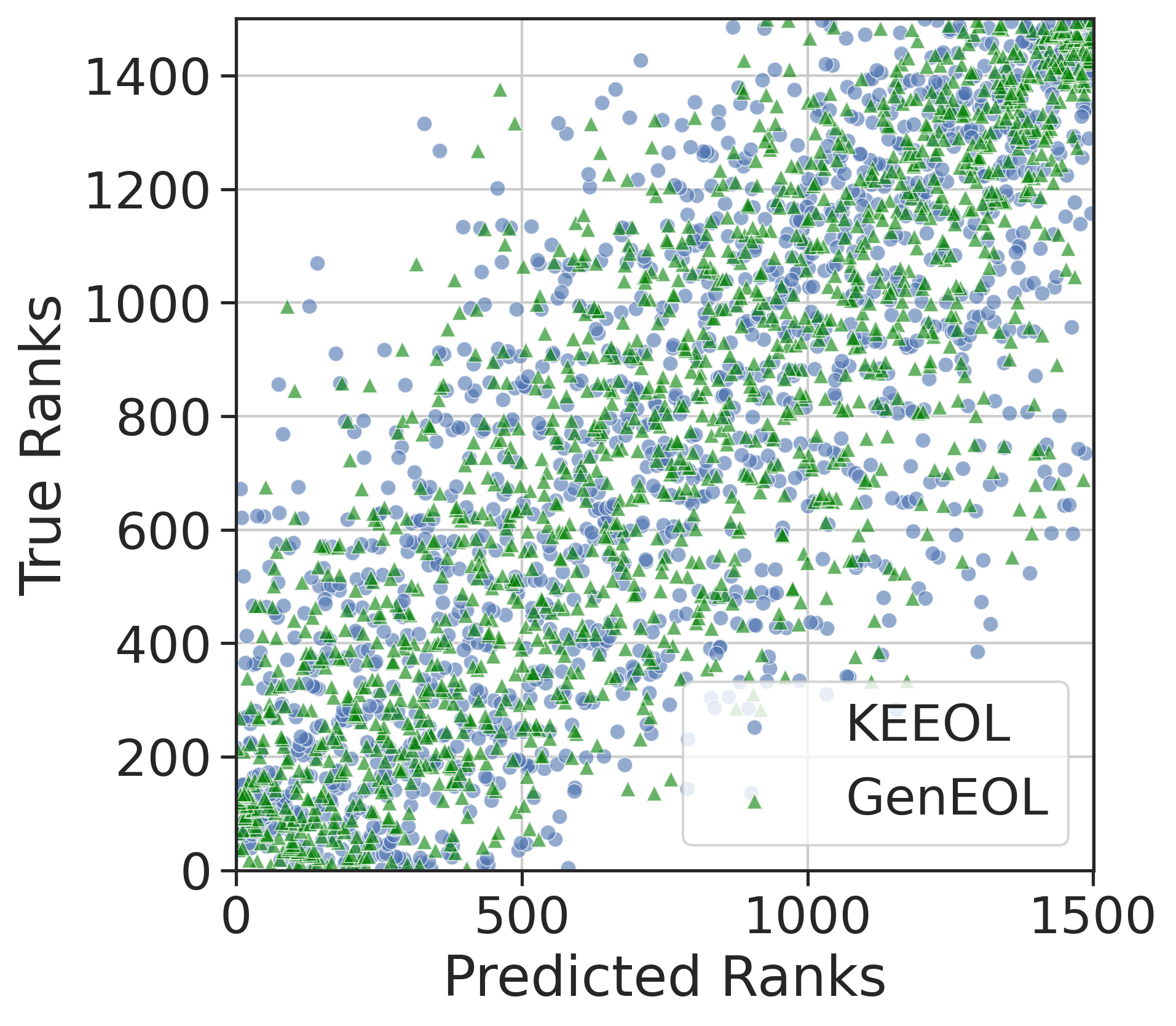}
        \caption{}
        \label{fig:obs2_2}
    \end{subfigure}
    \vspace{-0.7em}
    \caption{(a) Comparison of \textit{Pretrained} vs the \textit{Instruction Tuned} Models on STSB validation set. Pretrained models are always better in sentence embeddings. (b) Scatter Plot for score ranks on the STSB validation set using EOL based prompt. (ordinal ranks used for better visualization) (c) Scatter Plot for score ranks with \ours. The points are more concentrated along the diagonal compared to 
    \keeol\ (Note the blue dots away from the diagonal in (c)).
    }
    \vspace{-1.5em}
\end{figure*}

\noindent\textbf{Training Free Methods}: 
Echo \cite{springer2024repetition} showed that repeating the original sentence and using the hiddden representations of the later as the embedding of the sentence can improve performance. \prompteol\ \cite{jiang2023scaling} showed that generative LLMs when used with the prompt (\textit{This sentence: "[TEXT]" means in one word:"}) and using the hidden representation of the last token i.e. \textit{:"} is very effective at embedding the text. Further improving this prompt \citet{zhang2024simple} proposed two advanced prompts - 1. Pretended Chain of Thought prompt (\pcoteol) - (\textit{After thinking step by step , this sentence : "[X]" means in one word:"}) and 2. Knowledge Enhancement prompt (\keeol) - (\textit{The essence of a sentence is often captured by its main subjects and actions...."[X]" means in one word:"}). \metaeol\ \cite{lei2024meta} utilized eight diverse prompts from four categories—Text Classification, Sentiment Analysis, Paraphrase Identification, and Information Extraction—to aggregate multiple perspectives into a single final embedding. These methods also discuss better ways to extract text embeddings by using LLM penultimate layers. However, they do not fully leverage the generative power of LLMs for embedding sentences.

\section{Motivation and Methodology}
\subsection{Motivation}\label{sec:3.3}
This section outlines the STS sentence similarity task, presents key empirical observations that motivate our approach, and finally elaborates the proposed \geneol\ method.


\noindent\textbf{Task Definition}: Given a set of sentence pairs, $\{...(x_{i1}, x_{i2})...\}$, the goal is to generate embeddings for each sentence such that, when ranked by the cosine similarity of their embeddings, the ranking aligns with a provided reference ranking. Performance is evaluated using Spearman's rank correlation between the predicted ranks and the true ranks. The task assumes access to a pretrained LLM $\mathcal{L}_{PT}$ and an instruction-tuned LLM $\mathcal{L}_{IT}$.

\noindent\textbf{Observation 1: Instruction tuned LLMs are worse than Pretrained LLMs at Embeddings}
We first analyze in Fig. \ref{fig:obs1}, the performance (spearman rank correlation) of instruction-tuned LLMs versus pretrained models on the STSB validation set using the \keeol\ prompt \cite{zhang2024simple} for sentence encoding. Results indicate that pretrained models consistently outperform instruction-tuned models in embeddings. Additionally, when applying an instruction version of the knowledge EOL prompt (with \texttt{[INST]} tokens) to instruction-tuned models, performance decreased further. This may explain why recent studies \citet{zhang2024simple} and \citet{lei2024meta} benchmark non-CL training-free methods primarily using pretrained models. 
On the other hand, instruction tuned models are generally vastly superior at generating texts suited to various prompts. This motivates an investigation into the potential of leveraging the generative capabilities of an instruction-tuned model $\mathcal{L}_{IT}$  to enhance embeddings derived from a pretrained model $\mathcal{L}_{PT}$.


\noindent\textbf{Observation 2: LLM embedding scores are coarsely aligned with True scores.}
To visualize the rank correlation discussed above, we present a scatter plot comparing the true and predicted ranks for the \msone\ embedding model in Fig. \ref{fig:obs2}. An ideal such scatterplot would have all elements aligned along the diagonal. Despite the notable amount of spread in Fig. \ref{fig:obs2}, predicted scores are still coarsely aligned with the true scores.   

To reduce the spread around the diagonal, we consider the following exercise. Let $(x_{i1}, x_{i2})$ be the i-th datapoint. Let $\tau_i$ be the true score and $\rho_i$ be the predicted  similarity score of this datapoint. Considering $\rho_i$ to be a random variable, the coarse alignment trend from Fig \ref{fig:obs2} gives a sense of an error being present in $\rho_i$. Specifically, $\rho_i = \tau_i + \epsilon_i$, where $\epsilon_i$ is the error term with a mean $\mu_i$ and variance $\sigma^2_i$. From a statistics standpoint, the primary task now is to the reduce this error to improve performance. $\mu_i$ here is the inherent bias and would require further tuning the model to reduce it. To reduce the variance $\sigma^2_i$, a simple trick now would be to average $k$ independent estimates of $\rho_i$ i.e. $\{\rho^{1}_i, ..,\rho^{k}_i\}$. It is known that the variance of the mean of $k$ independent random variables decreases by a factor of $k$ and hence reduces the error $\epsilon_i$.

As we've seen in Observation 1, instruction tuned models although bad embedding models are very good at generating text of preferred format. Hence, we hypothesize that instruction tuned models can be prompted appropriately to transform datapoint $(x_{i1}, x_{i2})$ to $\{(x^0_{i1}, x^0_{i2}), (x^1_{i1}, x^1_{i2}),.., (x^k_{i1}, x^k_{i2})\}$ with each pair containing the exact same meaning i.e. having the exact same $\tau_i$. While these pairs may not be entirely independent, it would still contribute to reducing variance. Let's say we transform each sentence $m$ times and use their mean embedding to represent the sentence $x_{i1}$, i.e., $\sum_{j=0}^m\frac{h(x^{(j)}_{i1})}{m}$. This would result in $k=\sim m^2$ pairs in estimating $\rho^{mean}_i = \left<\left(\sum_{j=0}^m\frac{h(x^{(j)}_{i1})}{m}\right), \left(\sum_{j=0}^m\frac{h(x^{(j)}_{i2})}{m}\right)\right>$ where $<.,.>$ is the dot product. Therefore, smaller values of $m$ might still work pretty well. Fig. \ref{fig:obs2_2} shows the scatter plot with \geneol\ ($m=8$). As shown, a notable number of blue dots that were initially spread far from the diagonal, moved closer to the diagonal with \geneol\ (green triangles).

We now curate appropriate prompts that can realise the $m$ transformed sentences while preserving meaning. We take inspiration from CL training data generation methods - \sumcse, \syncse\ in doing so. 


\begin{table}[tbp]
\centering
\resizebox{0.76\columnwidth}{!}{\begin{tabular}{lr}
\toprule
\textbf{Transformation}                              & \multicolumn{1}{l}{\textbf{STSB}} \\
\hline
Baseline                                             & 78.47                             \\\hline
Diverse Transformations                              & 82.21                             \\\hline
Diverse Transformations                              & \multirow{2}{*}{82.68}            \\
+ Composition Transformation                         &                                   \\
\bottomrule
\end{tabular}}
\caption{Diverse transformations significantly improve performance on the STSB validation set. Compositional summary transformation yields notable gains.}
\label{tab:meth1}
\vspace{-1.5em}
\end{table}

\begin{table*}[htb]
\centering
\resizebox{\textwidth}{!}{\begin{tabular}{l|c|c|ccccccc|c}
\toprule
\textbf{Method} &\textbf{\textit{Generator}} & $\boldsymbol{m}$ & \textbf{STS12} & \textbf{STS13} & \textbf{STS14} & \textbf{STS15} & \textbf{STS16} & \textbf{STSB} & \textbf{Sick-R} & \textbf{Avg.} \\
\midrule\hline
\multicolumn{9}{c}{\textbf{\textit{Embedder}}: \msone}\\\hline
\Avg$^\dag$ & \multirow{6}{*}{-}&\multirow{6}{*}{-}&  41.13 & 54.08 & 43.99 & 56.94 & 53.80 & 42.99 & 52.32 & 49.32\\
\echo$^\dag$ &&&  58.43 & 78.53 & 68.42 & 78.82 & 77.52 & 73.85 & 71.95 & 72.5 \\
\prompteol$^\dag$  &&& 63.08 & 78.58 & 69.40 & 77.92 & 79.01 & 75.77 & 69.47 & 73.32 \\
\pcoteol$^\dag$ &&& 66.45 & 82.04 & 72.24 & 77.93 & 79.36 & 76.66 & 71.06 & 75.11 \\
\keeol$^\dag$ &&& 66.33 & 81.52 & 71.73 & 77.53 & 77.99 & 74.09 & 74.02 & 73.89 \\
\metaeol$^\dag$ &&& 64.05 & 82.35 & 71.57 & 81.36 & 79.98 & 78.29 & 75.13 & 76.09 \\
\hline
\multirow{4}{*}{\ours} & \multirow{2}{*}{\misione} & 8& 69.60 & 83.66 & 77.99 & 82.46 & 81.93 & 81.67 & 76.44                 & \underline{79.11} (+3.02)\\
 && 32 & 72.13 & 84.65 & 79.48 & 82.83 & 82.44 & 81.56 & 76.83                 & \textbf{79.99} (+3.9)\\\cline{2-11}
 &\multirow{2}{*}{\texttt{ChatGPT}} & 8 & 70.10 & 83.92 & 78.73 & 82.41 & 81.96 & 80.37 & 77.49                 & \underline{79.28} (+3.19)\\
 & & 32 & 72.38 & 84.62 & 79.19 & 82.74 & 81.85 & 81.14 & 77.76 & \textbf{79.96} (+3.87)\\
\hline\hline
\multicolumn{9}{c}{\textbf{\textit{Embedder}}: \lmtwo}\\\hline
\Avg$^\dag$ & \multirow{6}{*}{-}&\multirow{6}{*}{-}&  35.49 & 53.15 & 40.12 & 55.35 & 53.26 & 42.10 & 49.96 & 47.06\\
\echo$^\dag$   &&& 52.40 & 72.40 & 61.24 & 72.67 & 73.51 & 65.73 & 64.39 & 66.05 \\
\prompteol$^\dag$  &&& 58.81 & 77.01 & 66.34 & 73.22 & 73.56 & 71.66 & 69.64 & 70.03 \\
\pcoteol$^\dag$ &&& 67.45 & 83.89 & 74.14 & 79.47 & 80.76 & 78.95 & 73.33 & 76.86 \\
\keeol$^\dag$ &&& 66.60 & 82.62 & 74.48 & 80.75 & 80.13 & 80.34 & 75.89 & 77.14 \\
\metaeol$^\dag$ &&& 64.16 & 81.61 & 73.09 & 81.11 & 79.84 & 77.96 & 74.46 & 75.96 \\\hline
\multirow{4}{*}{\ours} & \multirow{2}{*}{\misione} & 8& 67.37 & 82.48 & 76.50                 & 81.33 & 79.81 & 80.03 & 77.68                 & \underline{77.89} (+0.75)\\
 && 32 & 70.24 & 83.43 & 78.03 & 81.79 & 80.65 & 80.46 & 78.08                 & \textbf{78.95} (+1.81)\\\cline{2-11}
 &\multirow{2}{*}{\texttt{ChatGPT}} & 8 & 68.39 & 82.48 & 77.00                 & 82.05 & 80.28 & 79.20 & 78.38                 & \underline{78.25} (+1.11)\\
 & & 32 & 70.78 & 83.28 & 77.75                 & 82.10 & 80.45 & 79.83 & 78.71 & \textbf{78.99} (+1.85)\\\hline\hline
\multicolumn{9}{c}{\textbf{\textit{Embedder}}: \lmthree}\\\hline
\prompteol$^\dag$  & \multirow{4}{*}{-}&\multirow{4}{*}{-}& 60.88 & 78.57 & 68.18 & 76.75 & 77.16 & 72.83 & 68.94 & 71.90 \\
\pcoteol &&& 65.38 &	82.44 &	71.26 &	79.22 &	79.80 &	77.99 &	72.49 & 75.51\\
\keeol &&& 62.18 &	82.35 &	73.04 &	80.13 &	80.17 &	78.95 &	77.23 & 76.29\\
\metaeol$^\dag$ &&& 65.10 & 83.08 & 73.01 & 81.57 & 81.47 & 80.47 & 76.46 & 77.35\\\hline
\hline
\multirow{4}{*}{\ours} & \multirow{2}{*}{\misione} & 8& 68.71 & 84.24 & 77.95 & 82.13 & 82.44 & 81.80 & 78.74                 & \underline{79.43} (+2.08)\\
 && 32 & 71.39 & 85.13 & 79.63 & 82.44 & 82.79 & 82.16 & 79.03                 & \textbf{80.37} (+3.02)\\\cline{2-11}
 &\multirow{2}{*}{\texttt{ChatGPT}} & 8 & 68.63 & 84.03 & 78.44 & 81.98 & 82.57 & 80.45 & 79.12                 & \underline{79.32} (+1.97)\\
 & & 32 & 71.07 & 84.67 & 78.90 & 82.09 & 82.66 & 81.15 & 79.33 & \textbf{79.98} (+2.63)\\
\bottomrule
\end{tabular}}
\caption{Performance on 7 STS tasks. \ours\ outperforms every other training free method across different \textit{embedder}s. Two best models are \textbf{bold}, next two are \underline{underline}. (.) is gain on top of previous best (same \textit{embedder}).  \pcoteol, \keeol\ use penultimate layer. All other methods use last layer. $\dag$: Numbers reported in prior works.
}
\vspace{-1.5em}
\label{tab:main}
\end{table*}

\subsection{Methodology}\label{sec:3}

Figure \ref{fig:main} illustrates the methodology for the proposed approach. Our method, \geneol\ uses the LLMs in two distinct roles 1. \textbf{\textit{Generator} (uses $\mathbf{\mathcal{L}_{IT}}$)}; 2. \textbf{\textit{Embedder} (uses $\mathbf{\mathcal{L}_{PT}}$)}.
Although our motivation necessitates independent transformations of $x_i$, the method to achieve this remains uncertain. Therefore, we employ the \textit{generator} to facilitate diverse transformations of $x_i$. The transformations are accomplished by applying suitable prompts to the input sentence $x_i$ and utilizing LLM $\mathcal{L}_{IT}$ to generate the modified sentence.
\subsubsection{Diverse Transformations}
 \noindent\textbf{Transformation 0} ($\mathcal{T}_0$): Original sentence, $x_i$, is the only transformation that retains all aspects and meaning of the sentence.\\
\noindent\textbf{Transformation 1} ($\mathcal{T}_1$): Syntax of sentences has been shown to confuse sentence embeddings \cite{zhang2023well}. Hence, $\mathcal{T}_1$ is a sentence structure changing transformation.\\
\noindent\textbf{Transformation 2} ($\mathcal{T}_2$): Removing non essential details like adverbs shouldn't change the core meaning of the sentence. Hence, following \citet{zhang2023contrastive}, we use concise sentence transformation as $\mathcal{T}_2$.\\
\noindent\textbf{Transformation 3} ($\mathcal{T}_3$): Entailment is another transformation which has low semantic overlap while retaining the core meaning of a sentence \cite{gao2021simcse}. This becomes $\mathcal{T}_3$.\\
\noindent\textbf{Transformation 4} ($\mathcal{T}_4$): A regular paraphrasing that can retain the meaning is used as $\mathcal{T}_4$ \cite{zhang2023contrastive, thirukovalluru2024sumcse}.

Few shot prompts used for these transformations are detailed in \S \ref{ap:transformations}. An example of transformed sentences with these prompts is shown in Table \ref{tab:tansformed_sentences} (\S \ref{sec:appendix}). These prompts and demonstrations emphasize that the rephrased sentence must convey the same meaning as the original sentence. To evaluate the effectiveness of these diverse transformations, we analyze their impact on the STSB validation set. Table \ref{tab:meth1} shows that using diverse transformations to aggregate a sentence embedding is significantly more effective than just using the original sentence.

\subsubsection{Further Increasing Diversity (Optional)}
\sumcse\ \cite{thirukovalluru2024sumcse} showed that compositional summary transformations (a second summary transformation over the first transformation) is very effective at creating transformed sentences far from the original sentence. Compositional transformations retain the original meaning of the sentence while more significantly altering its lexical form compared to a single transformation. As shown in Table \ref{tab:meth1}, a compositional summary transformation (\S \ref{ap:transformations}) (emphasizes for meaning preservation in contrast to \sumcse\ prompt) shows benefits. Thus, we optionally incorporate the compositional summary transformation.

\subsubsection{Final Embedding}
The proposed framework, \ours, allows for use of other diverse transformations or sampling multiple diverse sentences ($m$ number of them) from the proposed transformations to improve sentence embeddings. The transformed sentences ($m$) and the original sentence ($1$) are embedded using the \textit{embedder}, LLM $\mathcal{L}_{PT}$ using the \keeol\ prompt \cite{zhang2024simple}. As shown in Fig. \ref{fig:main}, the final sentence embedding is the mean of the ($m+1$) embeddings.

\section{Experiments}
We evaluate \geneol\ and other baselines on the STS benchmark \cite{conneau2018senteval}. Spearman rank correlation (cosine similarity) is the main metric \cite{muennighoff2022mteb}. Training sets of the STS tasks are not used. We additionally asses \ours\ on 10 MTEB tasks across 4 categories (Classification, Clustering, Reranking and Pair Classification) using the appropriate metrics described in \citet{muennighoff2022mteb}.
\subsection{Implementation}
We explore multiple choices for both \textit{generator} and \textit{embedder}. For \textit{generator}, we try with both small and large models: \misione\ (Mistral-7B-Instruct-v0.1) and \texttt{ChatGPT} (gpt-3.5-turbo-0125). \textit{Embedder} is the common module that \ours\ shares with other baselines and becomes the basis of comparison. For \textit{embedder}, we try three pretrained models based on previous methods - \msone\ (Mistral-7B-v0.1), \lmtwo\ (Llama-2-7b-hf), \lmthree\ (Meta-Llama-3-8B).


\textit{Embedder} in \geneol, by default, uses the embedding prompt from \keeol\ \cite{zhang2024simple} i.e. \textit{The essence of a sentence is often captured by its main subjects and actions, while descriptive terms provide additional but less central details. With this in mind , this sentence : "[X]" means in one word:"}. We evaluate the performance of \geneol\ at different values of $m$. When $m>4$, multiple transformed sentences are sampled from the \textit{generator} from each transformation.

\subsection{Results STS}
Table \ref{tab:main} shows results of multiple methods on STS. Results for \keeol\ and \pcoteol\ use the penultimate layer as proposed in \citet{zhang2024simple}. Every other method (including \ours) uses the final layer. \ours\ significantly beats the next best method by \textbf{(3.88 , 1.83, 2.83)} points on average with (\texttt{Mistral0.1-7B}, \texttt{Llama2-7B}, \texttt{Llama3-8B}) respectively on the 7 STS task average. The average gain of \ours\ ($m=32$) is a \textbf{2.85 points} higher than the previous best method across models. \textbf{\geneol\ demonstrates consistent performance improvements across all \textit{embedder} models}, a trend that is uncommon in previous approaches.

\begin{table*}[htbp]
\centering
\resizebox{\textwidth}{!}{\begin{tabular}{l|c|ccc|cc|ccc|cc|c}
\toprule
\multicolumn{1}{l}{\textbf{Method}}  & \multicolumn{1}{|c}{\multirow{2}{*}{\textbf{\textit{Embedder}}}} & \multicolumn{3}{|c}{\textbf{Classification}}                                                    & \multicolumn{2}{|c}{\textbf{Clustering}}                        & \multicolumn{3}{|c}{\textbf{Reranking}}                                                   & \multicolumn{2}{|c|}{\textbf{Pair Classification}}        &    \multicolumn{1}{c}{\multirow{2}{*}{\textbf{Avg.}}}                       \\\cline{3-12}
                        &\multicolumn{1}{c}{}& \multicolumn{1}{|c}{\textbf{AC}} & \textbf{B7} & \multicolumn{1}{c}{\textbf{EC}} & \multicolumn{1}{|c}{\textbf{MX}} & \multicolumn{1}{c}{\textbf{TN}} & \multicolumn{1}{|c}{\textbf{AU}} & \textbf{SD} & \multicolumn{1}{c}{\textbf{SO}} & \multicolumn{1}{|c}{\textbf{TS}}  & \textbf{SD} & \\\midrule
\multicolumn{12}{c}{\textit{Unsupervised}}\\\midrule
\simcse\ & \texttt{Bert-Large} & 67.09 & 73.55 & 42.22 & 21.97 & 23.21 & 51.57 & 66.33 & 39.35 & 60.21 & 69.41 & 51.49 \\
\llmtovec\ & \msone &\textbf{76.94} & \textbf{86.16} & 48.88 & 26.93 & 30.26 & 58.60 & 77.81 & \textbf{49.80} & 68.76 & \textbf{91.30} & 61.54 \\\midrule
\multicolumn{12}{c}{\textit{Training-Free}}\\\hline
\prompteol & \multirow{6}{*}{\misione} & 71.91 & 76.70 & 45.75 & 26.52 & 36.27 & 55.85 & 78.54 & 43.00 & 67.25 & 32.21 & 53.40 \\
\pcoteol && 71.88 & 75.89 & 44.10 & 25.58 & 34.17 & 57.30 & 78.27 & 45.10 & 67.82 & 40.24 & 54.03 \\
\keeol && 72.72 & 78.70 & 48.46 & 24.54 & 30.48 & 53.95 & 73.44 & 39.12 & 71.03 & 43.70 & 53.61 \\ 
\metaeol && 74.85 & 83.26 & \textbf{52.75} & 28.48 & 41.16 & 57.21 & 80.68 & 43.27 & 68.76 & 72.49 & 60.29 \\
\tseol && 74.91 & 82.89 & 50.70 & 30.88 & 50.87 & 62.52 & 85.26 & 47.78 & 73.75 & 69.18 & 62.87 \\ 
\gentseol\ ($m=8$) && 71.59 & 82.73 & 47.54 & \textbf{31.14} & \textbf{51.78} & \textbf{63.58} & \textbf{85.44} & 48.45 & \textbf{76.82} & 77.58 & \textbf{63.67} \\ \bottomrule
\end{tabular}}
\caption{Results on MTEB tasks across 4 categories. \tseol\ is a task specific EOL prompts. \tseol\ does better than \keeol\ for all tasks. \gentseol\ (\textit{generator} = \texttt{ChatGPT}) does better than \tseol\ in all tasks except for the classification.}
\label{tb:ab6}
\vspace{-0.5em}
\end{table*}

For a more fair comparison, we additionally provide results with \ours\ at $m=8$ generations. \metaeol\ uses eight diverse prompts and averages 8 embeddings of the input sentence. \ours\ averages among $m+1=9$ embeddings. \ours\ ($m=8$) beats \metaeol\ by \textbf{2.41 points} on average across models. This shows that "\textbf{Diverse meaning retaining transformations can significantly improve sentence embeddings}".

\subsection{Results MTEB}\label{sec:mteb}
The STS tasks analysed so far are all based on sentence similarity, which is the main focus of this paper. In this subsection, we asses the performance of our method on other tasks from MTEB benchmark \cite{muennighoff2022mteb}. As our method is slightly expensive, we specifically pick tasks that are tractable for our method. We skip tasks that contain large number of datapoints (E.g. retrieval tasks have millions of documents). Details on specific tasks reviewed in each category are in \S \ref{ap:datasets}.

\keeol\ prompt was designed keeping the STS sentences in mind. It might not be suitable for other tasks like clustering medical paper titles. Hence, we first come up with appropriate EOL prompts for the different tasks mentioned. We call this collection - Task Specific EOL (\tseol) (\S \ref{ap:tsep_prompts}). \tseol\ outperforms \keeol\ across all MTEB tasks tried.

We use the same transformations as described in \S \ref{sec:3} to generate transformed sentences for the above tasks. We use \texttt{ChatGPT} as the \textit{generator} to asses the performance, minimizing any poor generations. These transformed generations are then used with \tseol\ prompts to get numbers for \gentseol. As shown in Table \ref{tb:ab6}, \textbf{\gentseol\ outperforms all other training-free methods on the MTEB tasks. It also beats strong unsupervised methods like \llmtovec\ \cite{behnamghader2024llm2vec} (which include contrastive training)}.

\gentseol\ results in notable gains across diverse tasks except for classification. Classification tasks often rely on specific aspects of a sentence (E.g. `Emotion' in EC and `Intent' in B7). These aspects are very specific to the original sentence. Transformations in \gentseol\ often change sentence structure, active speech to passive, etc. and might not retain such specific aspects like emotion. Hence, \gentseol\ performs worse on classification tasks. The proposed \gentseol\ framework allows for other transformations than ones prescribed in \S \ref{sec:3}. We believe such task specific transformations can further improve \gentseol. We leave this exploration for future work.

\subsection{Ablation 1: Effect of increasing the number of transformed sentences, \texorpdfstring{$\boldsymbol{m}$}{m}}
\begin{figure}[t]
    \vspace{-0.64em}
     \centering
     \includegraphics[width=0.93\columnwidth]{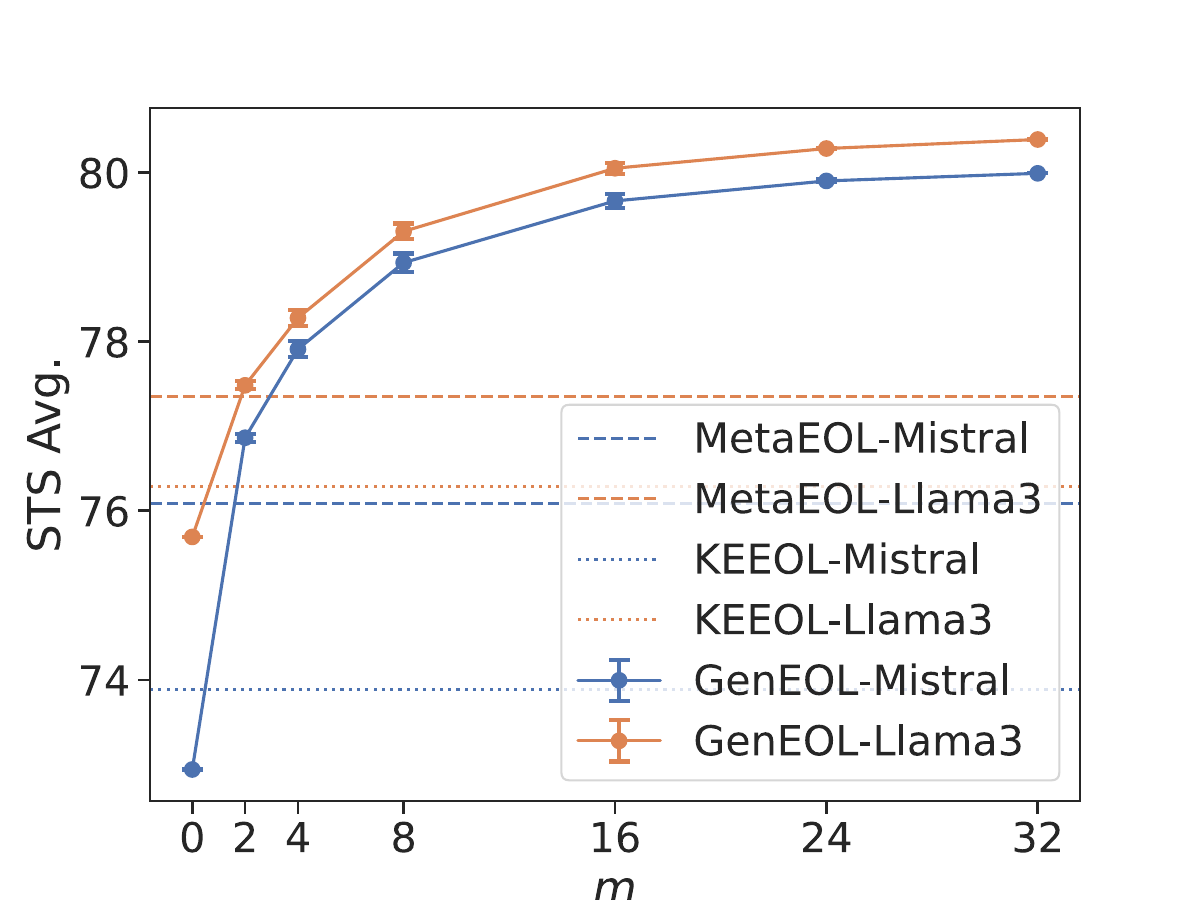}
    \caption{Average STS performance with $m$. \geneol\ beats \metaeol\ with only 2 transformations ($m$=2) for both \textit{embedder}s. Performance starts to stagnate around 16 generations. \geneol\ results averaged over 3 seeds.
    }
    \label{fig:ab2}
    \vspace{-1.5em}
\end{figure}

The number of transformed sentences, denoted as $m$, is a key parameter that significantly influences the performance of \geneol. To provide a clearer understanding of \geneol's effectiveness, we present results with varying $m$ values for both the \msone\ and \lmthree\ embedding models in Fig. \ref{fig:ab2}. For $m=2$, we randomly sample from transformations $\mathcal{T}_1$ and $\mathcal{T}_2$. For $m\geq4$, we maintain and equal diversity across all transformations - ($\mathcal{T}_1...\mathcal{T}_4$). \textbf{Even with just two generations, \geneol\ ($m=2$) beats all other baselines on STS.}

Increasing $m$ results in a significant increase in performance from $m=0$ to $m=32$ of over 5 points. The improvements are lower at higher $m$. Performance starts to stagnate after $m=16$. This is because similar transformed sentences/repetitions are sampled from the \textit{generator} at higher $m$ for a fixed number of transformations (4 in our case). Hence, \textbf{more diverse transformations might be required at higher $\boldsymbol{m}$ values}.

\begin{table*}[ht]
\centering
\resizebox{\textwidth}{!}{\begin{tabular}{l|ccccccc|c}
\toprule
\textbf{Method} & \textbf{STS12} & \textbf{STS13} & \textbf{STS14} & \textbf{STS15} & \textbf{STS16} & \textbf{STSB} & \textbf{Sick-R} & \textbf{Avg.} \\
\midrule\hline
\multicolumn{9}{c}{\textbf{\textit{Generator}}: \misione; \quad $m=8$; \quad \textbf{\textit{Embedder}}: \msone;}\\\hline
\ours\  & \textbf{69.60} & \textbf{83.66} & \textbf{77.99} & \textbf{82.46} & \textbf{81.93} & \textbf{81.67} & 76.44 & \textbf{79.11} \\\hline
\ours\ \texttt{w/o Composition}   & \underline{69.02} & \underline{83.30} & \underline{77.16} & \underline{81.61} & \underline{81.84} & \underline{80.55} & \underline{76.52} & \underline{78.57} \\\hline
\ours\ \texttt{w/o Composition, w/o $\mathcal{T}_2$, $\mathcal{T}_3$, $\mathcal{T}_4$} & 68.98 & 82.54 & 76.37 & 81.48 & 80.73 & 79.38 & \textbf{76.93} & 78.06 \\
\ours\ \texttt{w/o Composition, w/o $\mathcal{T}_1$, $\mathcal{T}_3$, $\mathcal{T}_4$} & 67.77 & 82.46 & 76.48 & 81.65 & 81.34 & 79.21 & 76.16 & 77.87\\
\ours\ \texttt{w/o Composition, w/o $\mathcal{T}_1$, $\mathcal{T}_2$, $\mathcal{T}_4$} & 61.69 & 81.27 & 74.55 & 78.23 & 78.21 & 72.96 & 73.22 & 74.30 \\
\ours\ \texttt{w/o Composition, w/o $\mathcal{T}_1$, $\mathcal{T}_2$, $\mathcal{T}_3$}  & 67.68 & 81.61 & 75.04 & 80.76 & 79.10 & 78.75 & 75.84 & 76.97 \\\hline
\geneol\ \texttt{w/o Composition, w/o $\mathcal{T}_1$, $\mathcal{T}_2$, $\mathcal{T}_3$, $\mathcal{T}_4$} & 60.25 &	78.69 &	69.60 &	76.79 &	76.54 &	75.21 & 73.50 &	72.94 \\
\bottomrule
\end{tabular}}
\caption{Dissecting \ours. $m=8$ for all rows. 1. Removing composition hurts performance (Row 2). 2. Non diverse transformations hurt performance (Rows 3-6). \ours\ with diverse transformations, composition is the best.}
\label{tab:ab1}
\vspace{-1.5em}
\end{table*}

\subsection{Ablation 2: Dissecting \geneol}
To understand the contribution of different components of \ours, we assess performance of individual components of \geneol. Table \ref{tab:ab1} shows results. All methods use the exact same value of $m=8$. Removing compositions results in notable reduction in  performance (-0.54). Hence, similar to \sumcse, summary compositions are important.

Among the individual transformations, $\mathcal{T}_1$ (changing sentence structure) performs the best. Using only transformed sentences from $\mathcal{T}_1$ results in a drop of 1.5 points. $\mathcal{T}_3$ (Entailment) performs the worst in this aspect resulting in a drop of 3.7 points. We posit this is because entailment transformation slightly changes the meaning in the transformed sentence. However, this still does better than not using any transformations.

\subsection{Ablation 3: Effect of Layer Number}
\begin{figure}[t]
    \vspace{-0.64em}
     \centering
     \includegraphics[width=0.95\columnwidth]{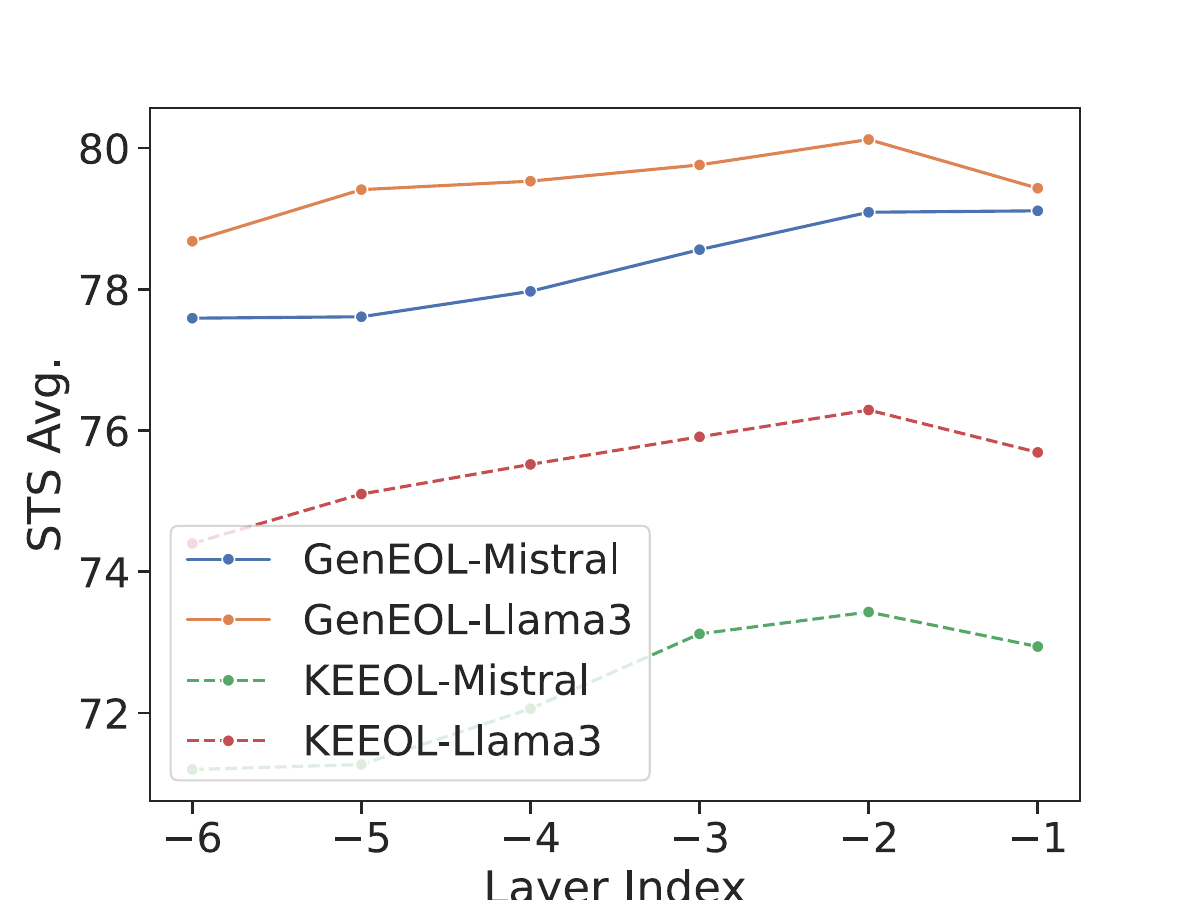}
    \caption{Avg. STS performance when embeddings are extracted from different layers. \geneol\ can  stabilize the representational quality across layers. Max-Min value across layers is lower for \geneol\ compared to \keeol}
    \label{fig:ab3}
    \vspace{-1.2em}
\end{figure}

Prior work on training free embeddings has shown that the last hidden layer might not be the most appropriate layer for text embeddings \cite{lei2024meta, zhang2024simple, li2024bellm}. We hence evaluate the performance of \geneol\ across different hidden layers. We perform the same analysis for \keeol. As shown in Fig. \ref{fig:ab3}, penultimate layer does best for \lmthree\ and second best for \msone. Performance drops beyond the penultimate layer.

Additionally, the variation (max-min) across layers is lower for \geneol\ at (1.52, 1.44) compared to \keeol\ at (2.22, 1.89) and \metaeol\ at (2.5, unknown) for (\msone, \lmthree) respectively. Hence, \textbf{\geneol\ can  stabilize the representational quality across the LLM layers}.

\subsection{Ablation 4: Compute Allocation in \geneol}
Datapoints in STS datasets comprise of a pair of sentences (\textit{sentence$_1$, sentence$_2$}) whose similarity needs to be predicted. Given a budget of generating 32 transformed sentences, \geneol\ by default invests this budget equally into both sentences in the pair. From the intuition discussed in \S \ref{sec:3.3}, this would create large pool of pairs which result in better averaging and improved results. Alternatively, one might invest this entire generation budget to \textit{sentence$_1$} or \textit{sentence$_2$} individually. 

In this ablation, we assess the effect of unequally splitting budget between the two sentences. Results are shown in Table \ref{tab:ab4}. Interestingly, allocating all generations to one of the sentences does worse than the baseline (without any generations i.e. $m=0$). This is counter-intuitive to \S \ref{sec:3.3}, which says that increasing the number of pairs improves performance. We posit this happens because a sentence when averaged with large number of embeddings, undergoes a normalizing effect and starts to focus on the core aspects of the sentence. On the other hand, the sentence which only has one embedding encodes some tail aspects about the sentence. This hurts the cosine similarity score between them. Including a few generations on both sentences performs significantly better i.e. the $(8,24)$ combinations perform much better than $(0,32)$ ones. Overall, \textbf{equally splitting the generation compute between the two sentences performs best} among the variants.

\begin{table}[t]
\centering
\resizebox{0.78\columnwidth}{!}{\begin{tabular}{l|cc|c}
\toprule
\textbf{Method} & \multicolumn{1}{c}{\textbf{$m_1$}} & \multicolumn{1}{c}{\textbf{$m_2$}} & \multicolumn{1}{|c}{\textbf{STS Avg.}} \\
\midrule
\ours            & 16                                 & 16                                 & \textbf{79.70}                                 \\
\ours\             & 24                                 & 8                                  & 79.30\\

\ours\           & 32                                 & 0                                  & 68.91                                 \\
\ours\             & 8                                 & 24                                  & 78.96\\
\ours\           & 0                                  & 32                                 & 69.92                                 \\
\keeol\ (Last Layer)           & -                                  & -                                  & 72.94 \\\bottomrule
\end{tabular}}
\caption{Performance of varying the compute on \textit{sentence$_1$} vs \textit{sentence$_2$}. Spending compute equally gives the highest gains. All compute spent on \textit{sentence$_1$} or \textit{sentence$_2$} does even worse than the baseline. }
\label{tab:ab4}
\vspace{-0.75em}
\end{table}

\subsection{Sensitivity to \textit{Embedder} Prompts}\label{sec:sep}
Changing prompts results in high variance for training free methods \cite{lei2024meta}. In this subsection, we asses the variance cause by using different prompts to the performance of training free embeddings. We compare our method to a baseline with varying prompts in Table \ref{tab:ab5}. Note all results are with the last hidden layer embeddings.

For the first experiment, we use the three major prompts discussed earlier - \prompteol, \pcoteol, \keeol. Results show that average variance of our method, \genseol, across models is lower than the baseline, \seol. For the second experiment, we make minimal changes to the \keeol\ prompt (removing/adding slack whitespaces) to form the \keeol$^{\prime}$ prompt shown in \S \ref{ap:spe}. \keeol$^{\prime}$ results in high variance for \seol, but stable performance (low variance) for \genseol. Thus, \textbf{\geneol\ is fairly robust to variations in embedding prompts}.

\subsection{Sensitivity to \textit{Generator} Prompts}
The prompts tried in the \S \ref{sec:3} were few shot prompts emphasizing that the core meaning needs to be retained. In this subsection, we try to understand the sensitivity of such \textit{generator} prompts. We vary both the \textit{generator} and the prompt to understand this. In the first experiment in Table \ref{tab:sgp}, we use the exact same prompts and change the \textit{generator}: we use \texttt{Llama-I-3.1} (Meta-Llama-3.1-8B). Although significantly better than the baseline from Table \ref{tab:main}, this result lags behind using \misione\ as the \textit{generator}.

For the second experiment in Table \ref{tab:sgp}, we merge the four few shot prompts and form a single prompt (without any demonstrations) and sample multiple times from it. As expected this drastically reduces performance with \misione\ due to low quality transformations. We perform a small qualitative analysis to understand the results.\\
\begin{table}[t]
\centering
\resizebox{\columnwidth}{!}{\begin{tabular}{l|c|cc}
\toprule
\textbf{Prompts} & \textbf{\textit{Embedder}} & \textbf{\genseol} & \textbf{\seol} \\\midrule
\multirowcell{2}[0pt][l]{\prompteol,\\ \pcoteol, \keeol} & \msone              &  \textbf{79.03 $\pm$ 0.88}       &  72.31	$\pm$ 0.76     \\
&\lmthree                 &    \textbf{79.76 $\pm$ 0.88}     &   73.42 $\pm$ 3.24  \\\hline
\multirow{2}{*}{\keeol, \keeol$^{\prime}$} & \msone    &  \textbf{79.98 $\pm$ 0.01}       &  71.92 $\pm$ 1.44     \\
&\lmthree                &    \textbf{80.17 $\pm$ 0.28}     &   74.48 $\pm$ 1.71  \\
\bottomrule
\end{tabular}}
\caption{(Avg. ± Std) across varying prompts. $*$ indicates prompt average. \genseol\ more stable than \seol.}
\label{tab:ab5}
\vspace{-1.2em}
\end{table}

\noindent\textbf{Qualitative Analysis}: Some examples for each of the four new runs from Table \ref{tab:sgp} are shown in \S \ref{ap:spg}. Poor quality generations that missed out essential details/hallucinated new content are highlighted. As can be seen, \texttt{Llama-I-3.1} with few shot prompts and \misione\ with zero shot prompt produced poor transformations. 

Overall, we suggest that tuning prompts on the development set could enhance \geneol\ for use with other \textit{generator} models and prompts. For instance, the compositional summary prompt in Table \ref{ap:tab:transformations} is derived from tuning its counterpart from \sumcse\ on the STSB validation set, resulting in an improvement in performance from 78.23 to 79.11 on the STS test set. Interestingly, \textbf{\texttt{ChatGPT} does very well with the single zero shot prompt to generate good quality transformed sentences}.

\begin{table}[ht]
\centering
\resizebox{0.92\columnwidth}{!}{\begin{tabular}{l|cc}
\toprule
\multicolumn{3}{c}{\quad \textbf{\textit{Embedder}}: \msone; \quad $m=8$;}\\\hline
\textbf{Prompt Type}                                                  & \textbf{\textit{Generator}}       & \textbf{STS Avg} \\\midrule
\multirow{3}{*}{Four Few Shot Prompts}                       & \misione & 79.11                       \\
\cline{2-3}
                                                             & \chatgpt & 79.28 \\
                                                             & \texttt{Llama-I-3.1}  & 77.48\\\hline
\multirow{2}{*}{Single Zero Shot Prompt} & \misione & 76.53                       \\
                                        & \texttt{ChatGPT}      & 79.89                       \\
\bottomrule
\end{tabular}}
\caption{Performance of various models under two distinct prompt configurations. \texttt{ChatGPT} demonstrates strong effectiveness even with a single zero-shot prompt.}
\label{tab:sgp}
\vspace{-1.5em}
\end{table}

\begin{table}[t]
    \centering
    \resizebox{0.95\columnwidth}{!}{\begin{tabular}{l|c|c}
        \toprule
        \textbf{Method} & \textbf{\textit{Embedder}} & \textbf{STS Avg.} \\
        \midrule
        \metaeol\ & \multirow{3}{*}{\misione} & 75.85 \\
        \geneol\ ($m=8$) && 79.11 \\
        \texttt{GenMetaEOL} ($m=8$) && \textbf{81.34} \\
        \bottomrule
    \end{tabular}}
    \caption{Integrating \geneol\ \textit{generator} transformations with \metaeol\ \textit{embedder} prompts yields significant gains, highlighting their complementary strengths.}
    \label{tab:genmetaeol}
    \vspace{-0.7em}
\end{table}

\subsection{Discussion: Difference from \metaeol\ and other training-free methods}
\geneol\ is a simple method that generates meaning-preserving transformations and aggregates them to improve training-free embeddings. While it may seem similar to other methods using multiple prompts, key differences exist.

For instance, \metaeol\ relies solely on prompting, while \geneol\ generates transformations compatible with any prompt.
\metaeol\ does not use any revised versions of input sentences. It uses eight \textit{embedder} prompts to get multiple embeddings capturing different aspects of a sentence (Eg. One prompt captures sentiment in it; One captures entities in the sentence; Another captures opinion; One assesses the type of the sentence, i.e., if it belongs to education/business/environment classes, etc.) \geneol\ uses a single \textit{embedder} prompt. It uses multiple \textit{generator} prompts to produce semantically equivalent transformations. To validate their synergy, we combine \geneol\ transformations with \metaeol\ prompts, introducing \texttt{GenMetaEOL}. Table \ref{tab:genmetaeol} shows that this combination yields significant performance gains justifying the complementary nature of the methods.

\texttt{QA-SimCSE} \cite{tu2024linguistically} is another method that generates intermediate text before generating embeddings for the given input. However, the work deals with a completely different task of conditional sentence similarity (An additional condition is used to assess similarity between two sentences) and it is not training-free.  \geneol\ produces $m$ meaning retaining transformations (as opposed to one conditional answer in \texttt{QA-SimCSE}) that are all used to embed the sentence.

\subsection{Discussion: Addressing the extra generation cost in  \geneol}\label{ss:extra_cost}
\geneol\ achieves superior performance over all training-free methods on the STS and MTEB benchmarks, demonstrating its effectiveness. However, this performance gain comes with increased computational cost. Table \ref{tab:runtime} provides a detailed comparison of the per example runtime for \textit{embedder}s and \textit{generator}s across methods, highlighting the higher overhead associated with \geneol.

\geneol's computational cost is higher due to generation of multiple transformations. However, this cost is incurred solely during inference, involving only forward passes, which are more readily parallelizable (vs training-intensive methods requiring backward passes) \cite{anil2018large, mccandlish2018empirical,thirukovalluru2024sequence}.

Recently, larger models like \texttt{GPT-4o} \cite{achiam2023gpt} and LLMonkeys \cite{brown2024large} have leveraged extra inference time compute to significantly improve reasoning tasks. Our work, \geneol, is the first such effort to use inference time compute to improve text embeddings. 

Moreover, techniques such as speculative decoding \cite{li2024eagle} and inference engines such as vllm \cite{kwon2023efficient} have made great strides in reducing the cost and latency of sampling from LLMs. Additionally, one might not need all 32 transformations to do a good job. One could adaptively scale compute for individual examples similar to MRL \cite{kusupati2022matryoshka}. All of these techniques can make \geneol\ more efficient.

\begin{table}[t]
    \centering
    \resizebox{0.78\columnwidth}{!}{\begin{tabular}{l|c|c}
        \toprule
        \textbf{Method} & \textbf{\textit{Embedder}} & \textbf{\textit{Generator}} \\
        \midrule
        \prompteol\ & 0.05 & 0.00 \\
        \pcoteol\ & 0.05 & 0.00 \\
        \metaeol\ & 0.24 & 0.00 \\
        \geneol\ ($m=8$) & 0.18 & 1.60 \\
        \geneol\ ($m=2$) & 0.08 & 0.40 \\
        \bottomrule
    \end{tabular}}
    \caption{Average per example runtime (in seconds) comparison on the STS. \geneol\ incurs higher cost. }
    \label{tab:runtime}
    \vspace{-0.7em}
\end{table}

\section{Conclusion}

We introduce \geneol, an effective method for leveraging the generative capabilities of large language models (including black-box models) to enhance sentence embeddings. Through extensive experiments and ablations, we show that aggregating embeddings of diverse meaning retaining transformations can significantly improve sentence embeddings. Even with very small number of transformed sentences i.e. ($m=2$), \geneol\ beats all baselines.  \geneol\ stabilizes representational quality across LLM layers and is robust to perturbations of embedding prompt. Despite gains on diverse MTEB tasks, \geneol\ falls short on classification tasks requiring identification of specific aspects about original sentence, such as emotion, subjectivity, or intent.

\section{Limitations}
The main limitation of \geneol\ is the cost associated with generating multiple transformations and then embedding them. Section \S \ref{ss:extra_cost} discusses a multiple techniques like parallel inference, inference engines, speculative decoding, adaptive compute scaling to address this issue and make \geneol\ computationally efficient.\\

\noindent\textbf{Broader Impact and Discussion of Ethics}:\\
While our model is not tied to any specific applications, it could be used in
sensitive contexts such as health-care, etc.  Any work using our
method is requested to undertake extensive quality-assurance and
robustness testing before applying in their setting. To the best of our knowledge, the datasets used in our work do not contain any sensitive information. \\

\noindent\textbf{License}: All datasets, methods used fall under Apache License 2.0. This research work abides by terms of the license. Research output of this paper also falls under Apache License 2.0.\\

\noindent\textbf{Replicability}:\\
Sourcecode: \href{https://github.com/raghavlite/GenEOL}{https://github.com/raghavlite/GenEOL}

\bibliography{custom}
\appendix

\section{Appendix}
\label{sec:appendix}

\subsection{Models and Parameters}\label{ap:training_params}
Model names and parameter counts are as follows.
\textbf{Generation Models Used}: \misione\ (Mistral-7B-Instruct-v0.1); \texttt{ChatGPT} (gpt-3.5-turbo-0125); \texttt{Llama-I-3.1} (Meta-Llama-3.1-8B)\\ 
\textbf{\textit{Embedder} Models Used}: \msone\ (Mistral-7B-v0.1), \lmtwo\ (Llama-2-7b-hf), \lmthree\ (Meta-Llama-3-8B).

\subsection{Compute}
All experiments were run on four A6000 GPUs (48gb). Generating transformations for each of the 7 STS datasets using \misione\ took 3 hrs for \geneol\ ($m=8$) and 12 hrs for ($m=32$). \chatgpt\ timings were approximately same.

\subsection{Datasets}\label{ap:datasets}

\noindent\textbf{MTEB}: We pick the following tasks-\\
\noindent\textbf{1. Classification}: AmazonCounterFactualClassification (AC), Banking77Classification (B7), EmotionCLassification (EC).\\
\noindent\textbf{2. Clustering}: MedarxivP2P (MX), TwentyNewsClustering (TN).\\
\textbf{3. Reranking}: AskUbuntuDupQuestions (AU), SciDocsRR (SD), StackOverflowDupQuestions (SO).\\
\textbf{4. Pair Classification}: TwitterSemEval2015 (TS), SprintDuplicateQuestions (SD).

\noindent Dataset sizes: \citet{conneau2018senteval, muennighoff2022mteb}

\subsection{Transformation Prompts}\label{ap:transformations}
All transformation prompts used in the work are listed in Table \ref{ap:tab:transformations}. For few shot prompts, demonstrations would follow the given text.

\begin{table*}[ht]
    \centering
    \begin{tabular}{@{}m{3cm} | m{12cm}@{}}
        \toprule
        Task Code & Description \\ \midrule
        $\mathcal{T}_1$ & Rewrite the input sentence or phrase using different sentence structure and different words while preserving its original meaning. Please do not provide any alternative or reasoning or explanation. \\ \midrule
        
        $\mathcal{T}_2$ & Create a sentence or phrase that is also true, assuming the provided input sentence or phrase is true. Please do not provide any alternative or reasoning or explanation. \\ \midrule
        
        $\mathcal{T}_3$ & Provide a concise paraphrase of the input sentence or phrase, maintaining the core meaning while altering the words and sentence structure. Feel free to omit some of the non-essential details like adjectives or adverbs. Please do not provide any alternative or reasoning or explanation. \\ \midrule
        
        $\mathcal{T}_4$ & Paraphrase the input sentence or phrase, providing an alternative expression with the same meaning. Please do not provide any alternative or reasoning or explanation. \\\midrule
        \texttt{Composition Summary} & Summarize the input sentence while preserving the exact meaning of the sentence. Do not output any additional explanation. Only output the summary.\\\midrule
        \texttt{Single Zero Shot Prompt} & Make changes to the following sentence to output 10 diverse sentences. Following are some directions for the changes. \\& 1. Paraphrase the input sentence or phrase, providing an alternative expression with the same meaning. \\& 2. Rewrite the input sentence or phrase using different sentence structure and different words while preserving its original meaning. \\& 3. Create a sentence or phrase that is also true, assuming the provided input sentence or phrase is true. \\& 4. Provide a concise paraphrase of the input sentence or phrase, maintaining the core meaning while altering the words and sentence structure. Feel free to omit some of the non-essential details like adjectives or adverbs. \\&\\& With these possible transformations in mind, transform the following sentence. Output the 10 transformed sentences in 10 new lines. Do not output any explanation. Do not output anything else.\\
        \bottomrule
    \end{tabular}
    \caption{For Few shot Prompts for the four transformations. Compositional Summary prompt also mentioned. Few shot demonstrations provided in sourcecode. Single zero short prompt also mentioned.}
    \label{ap:tab:transformations}
\end{table*}

\begin{table*}[ht]
    \centering
    \begin{tabular}{@{}>{\centering\arraybackslash}m{3cm} m{12cm}@{}}
        \toprule
        \textbf{Transformation} & \textbf{Transformed Sentence} \\ \midrule
        Original Sentence  & Aloes, like most succulents, are very hardy and will withstand considerable neglect .\\\midrule
        \multirowcell{2}{$\mathcal{T}\_1$} & Like most succulents, aloes are very hardy and can tolerate considerable neglect.\\
& Like most succulents, aloes are very hardy and can withstand a considerable amount of neglect.
 \\ \midrule
        
        \multirowcell{2}{$\mathcal{T}\_2$} & Aloes, like many succulents, are very resilient and can tolerate a lot of neglect.\\
& Aloes, like most succulents, are hardy and can endure a lot of neglect. \\ \midrule
        
        \multirowcell{2}{$\mathcal{T}\_3$} & Succulents are very hardy and will withstand neglect.\\
& Aloes are very hardy and can tolerate considerable neglect.\\\midrule
        \multirowcell{2}{$\mathcal{T}\_4$} & Aloe plants, similar to other succulents, are extremely resilient and can tolerate a significant level of neglect.\\
& Aloes, like many succulents, are extremely resilient and can tolerate a great deal of neglect. \\
        
        \bottomrule
    \end{tabular}
    \caption{Examples of transformed sentences for each transformation with \misione. While the transformations themselves are not easily distinguitable, smaller nuances are present within the data. E.g. $\mathcal{T}\_1$ sentences have changed sentence structure; $\mathcal{T}\_3$ sentences are true statements based on original. }
    \label{tab:tansformed_sentences}
\end{table*}

\begin{table*}[htbp]
\resizebox{\textwidth}{!}{\begin{tabular}{l|l}
\toprule
\textbf{\textit{Generator}}                                                & \textbf{Transformations}                                                                                                                    \\\hline
Sentence                                        & Aloes, like most succulents, are very hardy and will withstand considerable neglect .                                                \\\hline
\multirowcell{8}{\texttt{ChatGPT}\\(Single Zero\\ Shot Prompt)} & 
Aloes, similar to many other succulents, can tolerate a lot of neglect.\\&
Aloes are extremely resilient plants, much like most succulents.\\&
Aloes are known for their ability to thrive even when neglected, a common trait among succulents.\\&
Most succulents, including aloes, are quite hardy and can survive neglect.\\&
Aloes, along with most succulents, have a high tolerance for neglect.\\&
Aloes, as with many succulents, are very resilient and can handle being neglected.\\&
Aloes, much like the majority of succulents, are hardy plants that can withstand neglect.\\&
Most succulents, such as aloes, are sturdy and can withstand neglect.\\\hline
\multirowcell{8}{\misione\\ (Four Few Shot\\ Prompts)}         & Aloe plants, similar to other succulents, are extremely resilient and can tolerate a significant level of neglect.\\&
Aloes, like many succulents, are extremely resilient and can tolerate a great deal of neglect.\\&
Like most succulents, aloes are very hardy and can tolerate considerable neglect.\\&
Like most succulents, aloes are very hardy and can withstand a considerable amount of neglect.\\&
\hl{Succulents are very hardy and will withstand neglect.}\\&
Aloes are very hardy and can tolerate considerable neglect.\\&
Aloes, like many succulents, are very resilient and can tolerate a lot of neglect.\\&
Aloes, like most succulents, are hardy and can endure a lot of neglect.                                         \\\hline
\multirowcell{8}{\texttt{Llama3.1-I-8B}\\ (Four Few Shot\\Prompts)}                & Like most succulents, aloes are extremely resilient and can tolerate significant neglect.\\&
Like most succulents, aloe plants are extremely resilient and can tolerate significant neglect.\\&
Succulents, including aloes, are known for their remarkable hardiness and ability to thrive even when neglected.\\&
Succulents, including aloes, are known for their remarkable hardiness and ability to thrive despite considerable neglect.\\&
Many plants can tolerate some level of neglect.\\&
\hl{Most succulents are very hardy plants.}\\&
\hl{Succulents, including aloes, are resilient plants.}\\&
\hl{Succulents like aloes are very resilient.}\\\hline
\multirowcell{10}{\misione \\(Single Zero\\ Shot Prompt)}             & Aloes are hardy like most succulents and can tolerate significant neglect.\\&
Most succulents, including aloes, are hardy and can withstand considerable neglect.\\&
Aloe vera, like most succulents, is very hardy and can endure considerable neglect.\\&
Like most succulents, aloes are extremely hardy and can withstand significant neglect.\\&
\hl{Aloe vera, being a succulent like most others, is very hardy and can tolerate considerable neglect.}\\&
Aloe vera, like the majority of succulents, is extremely hardy and can endure considerable neglect.\\&
Aloe vera is very hardy, like most succulents, and can withstand considerable neglect.\\&
Like most succulents, aloe vera is hardy and can endure significant neglect with ease.\\\bottomrule                           
\end{tabular}}
\caption{Some bad transformations are highlighted. They have one of the two problems - 1. Not all information about the original sentence is retained. 2. LLM hallucinates some content.  }
\label{ap:tab:spg}
\end{table*}

\subsection{Sensitivity to \textit{Embedder} Prompts}\label{ap:spe}
\S \ref{sec:sep} uses multiple prompts to check the sentitivity of \geneol\ to \textit{embedder} prompts. These prompts are listed in Table \ref{ap:tab:spe}.

\begin{table*}[htbp]
\resizebox{\textwidth}{!}{
\begin{tabular}{l|l}
\toprule
\textbf{Prompt Version}   & \textbf{Prompt Text}\\\midrule
\prompteol & This sentence : "\{input\_text\}" means in one word:"\\\hline
\pcoteol   & After thinking step by step , this sentence : "\{input\_text\}" means in one word:"\\\hline
\multirow{3}{*}{\keeol}     & \multirowcell{3}[0pt][l]{The essence of a sentence is often captured by its main subjects and actions,\\ while descriptive terms provide additional but less central details. With this in\\ mind , this sentence : "\{input\_text\}" means in one word:"}\\\\\\\hline
\multirow{3}{*}{\keeol$^\prime$}   & \multirowcell{3}[0pt][l]{The essence of a sentence is often captured by its main subjects and actions,\\ while descriptive terms provide additional but less central details. With this in\\ mind, this sentence: "\{input\_text\}" means in one word:"}\\ \\ \\ \bottomrule
\end{tabular}}
\caption{Table containing all prompts used in the work. \keeol and \keeol$^\prime$ differ only by two whitespaces.}
\label{ap:tab:spe}
\end{table*}

\subsection{Sensitivity to \textit{Generator} Prompts}\label{ap:spg}
Transformed sentences using the four LM prompt combinations are shown in Table \ref{ap:tab:spg}. Some poor transformed sentences that either do not retain all aspects from the original sentence or contain some extra content compared to the original sentence are highlighted. \texttt{Llama3.1-I-8B} with four few shot prompts and \misione\ have some poor quality transformed sentences.

\subsection{Task Specific Prompts for MTEB}\label{ap:tsep_prompts}
\S \ref{sec:mteb} tuned the EOL prompts to be more applicable to MTEB tasks. Thee tuned prompts collection namely \tseol\ is shown in Table \ref{ap:tab:tsep_prompts}.

\begin{table*}[ht]
    \centering
    \resizebox{\textwidth}{!}{\begin{tabular}{@{}>{\centering\arraybackslash}m{4cm} m{12cm}@{}}
        \toprule
        \textbf{Task Name} & \textbf{Description} \\ \midrule
        \multirowcell{2}{AmazonCounterfactual\\Classification} & In this task, classify a given Amazon customer review text as either counterfactual or not-counterfactual. For this task, this review: \texttt{"\{input\_text\}"} means in one word: \\ \midrule
        
        \multirowcell{2}{Banking77Classification} & In this task, given an online banking query, find the corresponding intents. For this task, this query: \texttt{"\{input\_text\}"} means in one word: \\ \midrule
        
        \multirowcell{2}{EmotionClassification} & In this task, classify the emotion expressed in the given Twitter message into one of the six emotions: anger, fear, joy, love, sadness, and surprise. For this task, this message: \texttt{"\{input\_text\}"} means in one word: \\ \midrule
        
        \multirowcell{2}{MedrxivClusteringS2S} & In this task, identify the main category of Medrxiv papers based on the titles. For this task, this title: \texttt{"\{input\_text\}"} means in one word: \\ \midrule
        
        \multirowcell{2}{TwentyNewsgroups\\Clustering} & In this task, identify the topic and theme of the news article. For this task, this article: \texttt{"\{input\_text\}"} means in one word: \\ \midrule
        
        \multirowcell{2}{AskUbuntuDupQuestions} & In this task, you need to retrieve duplicate questions from AskUbuntu forum. For this task, this question: \texttt{"\{input\_text\}"} means in one word: \\ \midrule
        
        \multirowcell{2}{SciDocsRR} & In this task, given a title of a scientific paper, retrieve the titles of other relevant papers. For this task, this title: \texttt{"\{input\_text\}"} means in one word: \\ \midrule
        
        \multirowcell{2}{StackOverflowDup\\Questions} & In this task, retrieve duplicate questions from StackOverflow forum. For this task, this question: \texttt{"\{input\_text\}"} means in one word: \\ \midrule
        
        \multirowcell{2}{TwitterSemEval2015} & In this task, retrieve tweets that are semantically similar to the given tweet. For this task, this tweet: \texttt{"\{input\_text\}"} means in one word: \\ \midrule
        
        \multirowcell{2}{SprintDuplicateQuestions} & In this task, retrieve duplicate questions from Sprint forum. For this task, this question: \texttt{"\{input\_text\}"} means in one word: \\ 
        \bottomrule
    \end{tabular}}
    \caption{\tseol\ prompts}
    \label{ap:tab:tsep_prompts}
\end{table*}

\end{document}